\documentclass{article}
\PassOptionsToPackage{numbers, compress}{natbib}

\usepackage[sglblindworkshop, final]{neurips_2025}
\workshoptitle{AI for Science
}
\usepackage{wrapfig}
\usepackage{adjustbox}
\usepackage{microtype}
\usepackage{graphicx}
\usepackage{subcaption}
\usepackage{booktabs} 

\usepackage{hyperref}
\usepackage{subcaption}
\usepackage{multirow}

\usepackage{algorithm}
\usepackage{algpseudocode}

\usepackage{amsmath}
\usepackage{amssymb}
\usepackage{mathtools}
\usepackage{amsthm}
\usepackage{enumerate}
\usepackage[capitalize,noabbrev]{cleveref}

\theoremstyle{plain}

\theoremstyle{definition}

\theoremstyle{remark}

\usepackage[textsize=tiny]{todonotes}

\title{	
Demystifying Protein Generation with \\Hierarchical Conditional Diffusion Models}

\author{
Zinan Ling\textsuperscript{1},
Yi Shi\textsuperscript{2},
Brett McKinney\textsuperscript{1},
Da Yan\textsuperscript{3},
Yang Zhou\textsuperscript{4},
Bo Hui\textsuperscript{1} \\
\textsuperscript{1}University of Tulsa \quad
\textsuperscript{2}Johns Hopkins University \\
\textsuperscript{3}Indiana University Bloomington
\quad
\textsuperscript{4}Auburn University\\
\texttt{bo-hui@utulsa.edu}
}

\begin{document}

\maketitle

\author{Anonymous}
\begin{abstract}
Generating novel and functional protein sequences is critical to a wide
range of applications in biology. Recent advancements in conditional diffusion models have shown impressive empirical performance in protein generation tasks. However, reliable generation of proteins remains an open research question in \textit{de novo} protein design, especially when it comes to conditional diffusion models. Considering  the biological function of a protein is determined
by multi-level structures, we propose a novel multi-level conditional diffusion model that integrates both sequence-based
and structure-based information for efficient end-to-end protein design guided by
specified functions. By generating representations at different levels simultaneously, our framework can effectively model the inherent hierarchical relations between different levels, resulting in an \textit{informative and
discriminative} representation of the generated protein. We also propose Protein-MMD (Maximum Mean Discrepancy), a new reliable evaluation metric, to evaluate the quality of generated protein with conditional diffusion models. Our new metric is able to capture both distributional and functional similarities between real and generated protein sequences while ensuring conditional consistency. Using conditional protein generation tasks with benchmark datasets, we demonstrate the efficacy of the proposed protein generation framework and evaluation metric.
\end{abstract}

\section{Introduction}
Designing proteins with specific biological functions is a fundamental yet formidable challenge in biotechnology. It benefits wide-ranging applications from synthetic biology to drug discovery \cite{watson2023novo,DBLP:conf/iclr/BoseAHFRLNKB024, huang2016coming,DBLP:conf/iclr/FengLJML24,DBLP:conf/iclr/Huang0ZZZZCW0Y24,DBLP:conf/iclr/Huang0ZZZZCW0Y24,lin2023evolutionary}. The challenge arises from the intricate interplay between protein sequence, structure, and function, which has not yet been fully understood \cite{dill2008protein}. Traditional methods, such as directed evolution, rely on labor-intensive trial-and-error approaches involving random mutations and selective pressures, making the process time-consuming and costly \cite{arnold1998design}. Recently, generative models have emerged as promising tools for protein design, enabling the exploration of vast sequence-structure-function landscapes \cite{ anand2022protein, fu2024latent, dauparas2022robust,DBLP:conf/iclr/TrippeYTBBBJ23}. However, existing generative models—including those focused on enzyme engineering, antibody creation, and therapeutic protein development—are typically task-specific and require retraining for new design objectives \cite{fu2024latent, dauparas2022robust}. These limitations impede their adaptability and scalability across different protein families.

While conditional generative models offer an end-to-end solution by directly linking the design process to the guidance, these models have been applied to protein generation~\cite{DBLP:conf/iclr/KomorowskaMDVLJ24,DBLP:conf/icml/KlarnerRMDT24,DBLP:conf/nips/GruverSFRHLRCW23}. In conditional protein generation tasks, maintaining conditional consistency across diverse contexts and ensuring functional relevance are critical \citep{trippe2022diffusion, hu2024cpdiffusion-ss}. Specifically, the generated proteins should fully adhere to the specified functional constraints \citep{gretton2012kernel}. At the same time, achieving diversity and novelty in generated proteins is essential for successful design. In the literature, structural novelty can be assessed using Foldseek \citep{van2022foldseek}, which performs rapid protein structure searches against databases like PDB \citep{berman2000protein} and AlphaFold \citep{jumper2021highly} to ensure the generated proteins are novel compared to known structures. Diversity is measured using TM-score \citep{zhang2005tm}, which calculates structural variation between the generated proteins themselves and between the generated and wild-type proteins \citep{hu2024cpdiffusion-ss}.

\begin{wrapfigure}{r}{0.6\linewidth}
    \centering
    \vspace{-11mm}
    \begin{subfigure}{0.235\linewidth}
        \includegraphics[width=\linewidth]{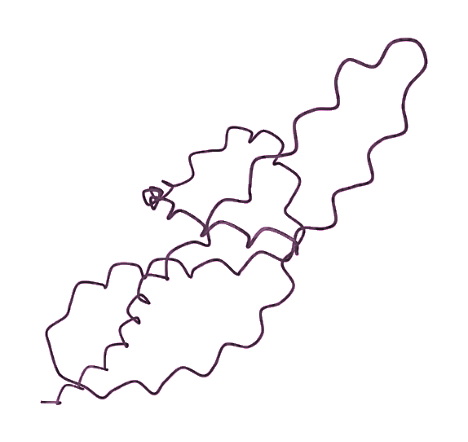}
        \caption*{MPNN}
    \end{subfigure}
    \hfill
    \begin{subfigure}{0.235\linewidth}
        \includegraphics[trim={0 0 0 5mm},clip,width=\linewidth]{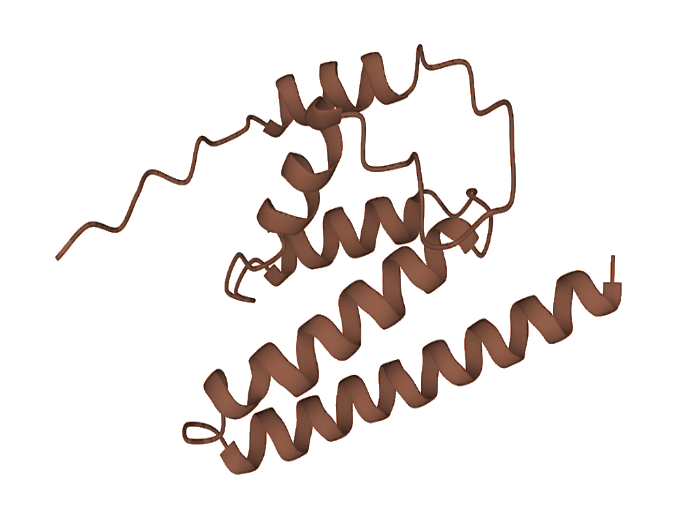}
        \caption*{ESM}
    \end{subfigure}
    \hfill
    \begin{subfigure}{0.235\linewidth}
        \includegraphics[width=\linewidth]{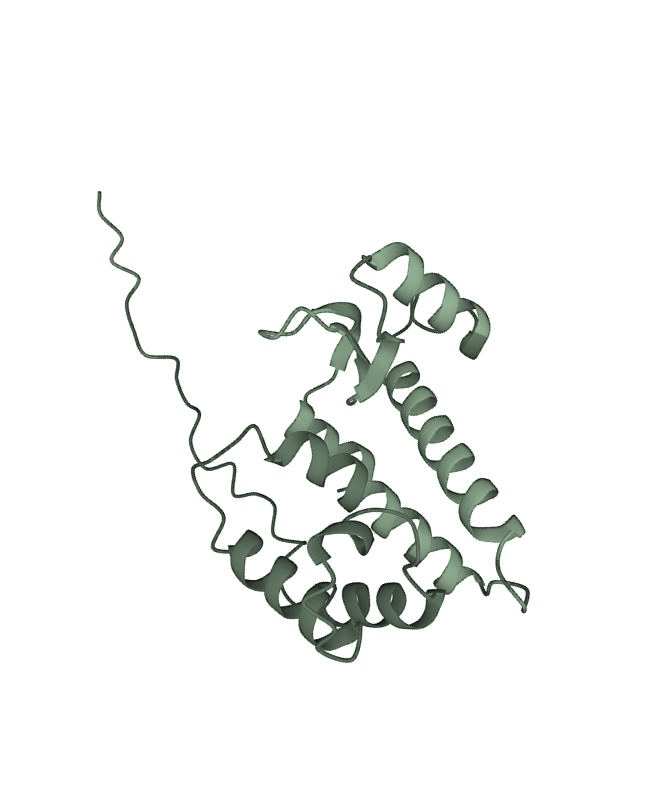}
        \caption*{GAN}
    \end{subfigure}
    \hfill
    \begin{subfigure}{0.235\linewidth}
        \includegraphics[width=\linewidth]{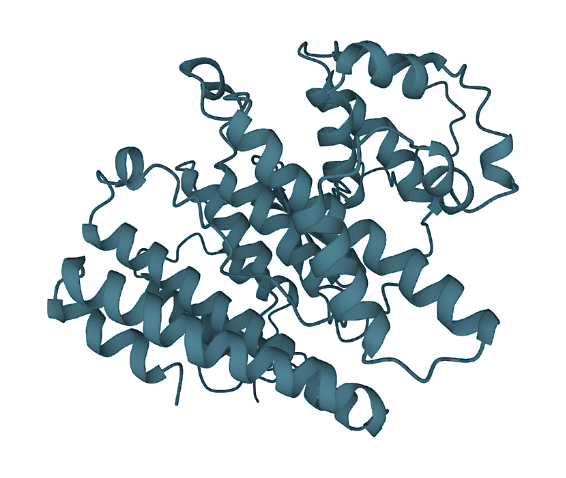}
        \caption*{Ours}
    \end{subfigure}
    \vspace{0mm}
    \caption{Protein visualization.}\label{fig:vote}
    \vspace{-3mm}
\label{figure: sample}
\end{wrapfigure}
Despite the success of existing diffusion models in protein generation, these models only  generate the protein representation at a single level and ignore hierarchical relations among
different levels of representations. Choosing the level of granularity at which representing the comprehensive information of the protein raises significant concerns about the reliability of generated proteins in real-world applications. Motivated by the need to capture both the structural and functional nuances of protein sequences, we propose a novel multi-level conditional generative diffusion model for protein design that integrates both sequence-based \citep{lin2023evolutionary} and structure-based \citep{wang2022learning} hierarchical information. Specifically, our
proposed method generates the protein at three different
levels: the amino acid level, the backbone level, and the all-atom level. Generation at multi-levels enables efficient end-to-end generation of proteins with specified functions and modeling the inherent hierarchical relations between different representations, resulting in an informative and
discriminative representation of the protein. Also, the conditional diffusion flow in the architecture preserves the hierarchical relations between different levels. Intuitively, a representation at the lower level (e.g., the atom level) can decide the potential representation space at the higher level (e.g., the amino acid level). Modeling such hierarchical relations can guarantee consistency at different levels. Our model incorporates a rigid-body 3D rotation-invariant preprocessing step combined with an autoregressive decoder to maintain SE(3)-invariance, ensuring accurate modeling of protein structures in 3D space. Figure~\ref{figure: sample} shows the proteins generated by different methods with the same input. The thin line indicates that the sequence is unlikely to undergo meaningful folding into a stable 3D structure. Compared with the baselines, our method can generate discriminative and functional proteins.

We remark that it is still unknown how to assess the \textit{conditional consistency}~\citep{gretton2012kernel} in \textit{de novo} protein design. Specifically, the fundamental problem of properly evaluating conditional consistency is quantifying to what extent the generated protein adheres to the specified functional constraints. Unlike computer vision, where metrics such as FID \citep{heusel2017gans} have become a standard for assessing generated images, it is unclear whether such metrics are suitable for protein generation tasks. In protein design, the generated output cannot be as easily visualized or assessed as in images, making the choice of evaluation metrics even more critical. Therefore, how to adapt metrics like FID or Maximum Mean Discrepancy \citep{gretton2012kernel} presents challenges. To address the challenges of evaluating the \textit{conditional consistency}, we propose \textit{Protein-MMD}, a metric based on  Maximum Mean Discrepancy (MMD), to better capture both distributional and functional similarities between real and generated protein sequences, while ensuring conditional consistency. We prove that our Protein-MMD provides a more accurate measure that reflects the given condition.
Experiments demonstrate that our proposed model outperforms existing approaches in generating diverse, novel, and functionally relevant proteins.
Our main contributions are summarized as follows:
\begin{itemize}
\vspace{-1mm}
    \item We design a novel multi-level conditional generative diffusion model that integrates sequence-based and structure-based information for efficient end-to-end protein design.
    \item We highlight the limitations of current evaluation metrics in protein generation and propose \textit{Protein-MMD}, a novel metric to evaluate conditional consistency for protein generation.
    \item We experiment with standard datasets to verify the effectiveness of the proposed model. Our evaluation metric paves the way for reliable protein design with given conditions.
\end{itemize}
\vspace{-2mm}

\vspace{-1mm}
\section{Methodology}
\subsection{Multi-level Diffusion}
Motivated by the need to capture both the structural and functional nuances of protein sequences, we propose a multi-level diffusion model to generate information about a protein at three levels: the amino acid level, the backbone level, and the all-atom level. By constructing representations at different levels, our framework effectively integrates the inherent
hierarchical relations of proteins, resulting in a more rational protein generative model. We remark that there are hierarchical relations among
different levels. To the best of our knowledge, this work is the first diffusion model
to generate information at three levels and leverage the hierarchical relation between different levels.

\begin{figure*}[h]
\vspace{-3mm}
  \includegraphics[width=\textwidth]{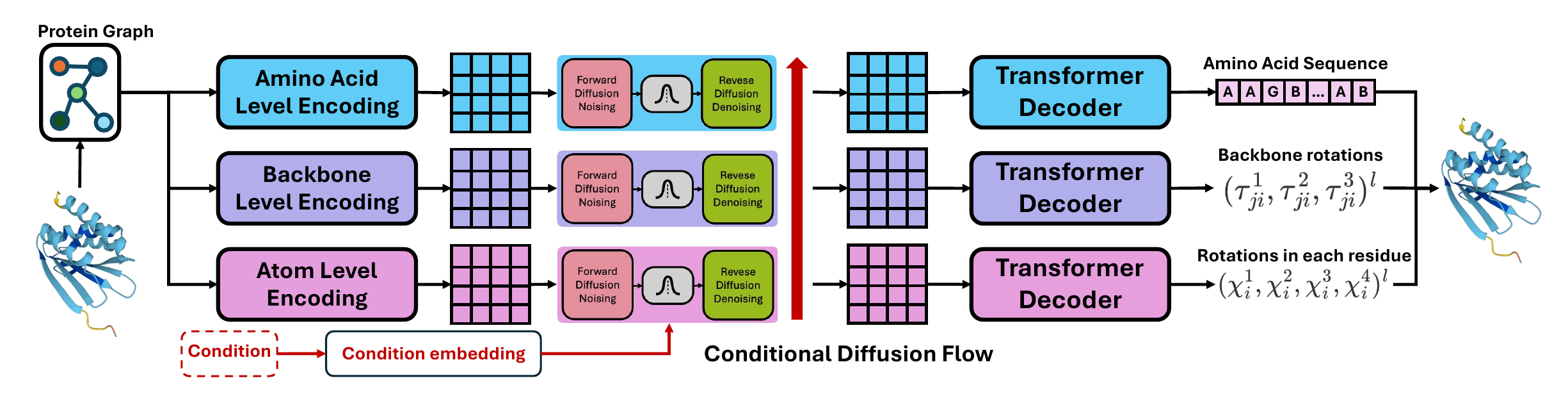}
  \vspace{-6mm}
  \caption{The architecture of the multi-level diffusion model.}
  \vspace{-3mm}
  \label{figure:arc}
\end{figure*}
 Figure~\ref{figure:arc} shows the architecture of our model. At each level, the information will be encoded with its own set of embeddings and processed through a conditional diffusion flow where the condition comes from a lower level. With decoders, the sequence, backbone rotations, and residue rotations will be combined to indicate the complete information of a generated protein.
 
 \vspace{1mm}
\noindent\textbf{Amino Acid Level Representation.} As the 3D conformation dictates biochemical interactions \citep{huang2016coming, dill2008protein}, we first represent a protein's structure as a graph $\mathcal{G}_a = (\mathcal{V}_a, \mathcal{E}_a)$, where $\mathcal{V}_a$ is the set of nodes corresponding to residues (amino acids), and $\mathcal{E}_a$ is the set of edges representing interactions between two residues. Specifically, an edge between two nodes $v_i$ and $v_j$ is established if the Euclidean distance between their $\text{C}\alpha$ atoms in 3D space is below a certain threshold, indicating a potential biochemical or structural interaction. At the amino acid level, each node $v_i \in \mathcal{V}_a$ corresponds to an amino acid and is represented by a vector $v_i = (\phi_i; h_i)$, where $\phi_i \in \mathbb{R}^3$ denotes the spatial coordinates of the amino acid's $\text{C}\alpha$ atom in three-dimensional space, and $h_i$ abstracts biochemical or structural properties. Each edge is represented as an embedding of the sequential distance~\cite{DBLP:conf/iclr/ZhangXJCLD023}.



 \vspace{1mm}
\noindent\textbf{Backbone Level Representation.}  An amino acid consists of backbone atoms and side chain atoms. Similarly, we use backbone atom ($C$, $N$, $C_\alpha$) coordinates as the feature of in node of the backbone $\mathcal{V}_b$.  We follow~\cite{DBLP:conf/iclr/ZhangXJCLD023} to compute three Euler angles $\tau_{i,j}^1$, $\tau_{i,j}^2$, $\tau_{i,j}^3$ between two backbone atoms $i$ and $j$. The angles will be integrated with the sequential distance as the edge feature.
Backbone-level representation derives finer-grained protein information. With the three angles, the orientation between any two backbone planes can be
determined to capture the backbone structures.

 \vspace{1mm}
\noindent\textbf{Atom Level Representation.} Atom-level representation considers all atoms in the protein and provides the most fine-grained information. There are several methods to treat an atom as a node in the representation~\cite{DBLP:conf/iclr/HermosillaSLFVK21,DBLP:journals/corr/abs-2106-03843}. Side chain torsion angles are important properties of protein structures~\cite{jumper2021highly}. In this paper, we also consider geometric representation at the atom level by incorporating the first four torsion
angles: $\chi_i^1$,  $\chi_i^2$,  $\chi_i^3$, and  $\chi_i^4$. With the complete geometric representation at the atom level, the diffusion model can
capture 3D information about all atoms in a protein and distinguish any two
distinct protein structures in nature.

 \vspace{1mm}
\noindent\textbf{Encoding.} We adopt a graph neural networks model~\cite{DBLP:conf/nips/WangLLLJ22} to encode the representing at different levels by leveraging the message-passing mechanism. 
In many models dealing with the spatial positions of amino acids, SE(3)-equivariance is often leveraged to ensure the invariance of operations such as translation and rotation~\cite{DBLP:conf/iclr/BoseAHFRLNKB024,DBLP:conf/icml/YimTBMDBJ23}. We also introduce a novel method to ensure SE(3)-invariance by transforming each amino acid’s coordinates $\phi$. This step is crucial for facilitating the subsequent autoregressive decoding.

Given a protein chain, we first translate the coordinates such that the position of the first amino acid is moved to the origin, i.e., $(0, 0, 0)$. Then, we apply a rotation matrix to align the position of the second amino acid onto the positive $x$-axis:
\begin{equation}
    {R}_1 = {I} + \sin(\theta) {K} + (1 - \cos(\theta)) {K}^2,
\end{equation}
where $\theta$ is the rotation angle between a node $v$ and the $x$-axis, and ${K}$ is the skew-symmetric matrix derived from the cross-product of $v$ and the unit vector along the $x$-axis. The third amino acid is rotated around the $x$-axis to place it in the positive $xy$-plane:\vspace{-2mm}
\begin{equation}
    R_2 = \begin{pmatrix}
    1 & 0 & 0 \\
    0 & \cos(\psi) & -\sin(\psi) \\
    0 & \sin(\psi) & \cos(\psi)
    \end{pmatrix},
\end{equation}
where $\psi$ is the angle that brings the third amino acid into the $xy$-plane. This process is iteratively applied to all amino acids in the protein chain.

In each iteration, the next amino acid is positioned relative to the previous ones, aligning the structure step by step while preserving the overall 3D conformation. The method ensures that all residues maintain SE(3)-invariance, making the transformations consistent across the entire protein chain.





The decoder at each level is an autoregressive Transformer \citep{vaswani2017attention} model that reconstructs the protein at each respective level. The autoregressive decoder can then use these transformed embeddings to reconstruct the information of a protein. At the sequence level, the decoder predicts the next amino acid token in the sequence. At the backbone level and the atom level, the decoder predicts geometric features (e.g., bond angles and distances) in an autoregressive fashion of each amino acid in the protein chain. Our method facilitates the use of SE(3)-invariant embeddings within an autoregressive framework. The decoder's autoregressive nature allows it to progressively predict amino acid positions by leveraging the SE(3)-invariant representation.

\vspace{1mm}
\noindent\textbf{Proof of SE(3)- invariance of the Transformation. }
Let $\{\phi_i\}_{i=1}^n$ be the original coordinates of the amino acids in the protein chain. Consider an arbitrary rotation $R \in \mathrm{SO}(3)$ and translation $\Gamma \in \mathbb{R}^3$ applied to the protein, resulting in transformed coordinates:
\begin{equation}
    \phi'_i = R \phi_i + \Gamma.
\end{equation}

Our goal is to show that after applying the transformation method to both $\{\phi_i\}$ and $\{\phi'_i\}$, the resulting representations are identical.

\noindent{\textit{Proof}}: For any transformation $T$ in $\mathrm{SO}(3)$ and any vector $v \in \mathbb{R}^3$, we have:
\begin{equation}
    T(v) = R v.
\end{equation}
Since rotations preserve vector norms, we can express $T(v)$ in terms of the norm of $v$ and its unit vector $v' = v / \|v\|$:
\begin{equation}
    T(v) = \|v\| R v' = \|v\| T(v').
\end{equation}
This implies that the effect of $T$ on $v$ can be decomposed into scaling by $\|v\|$ and transforming its direction via rotation and translation. To simplify the expression and subsequent calculations, we denote all vectors $\phi_i$ as unit vectors (i.e., their norms are equal to 1).

\textit{Step 1:} Translation to Origin
Compute the relative positions with respect to the first amino acid:
\begin{align}
    \xi_i &= \phi_i - \phi_1, \\
   \xi'_i &= \phi'_i - \phi'_1 = (R \phi_i + \Gamma) - (R \phi_1 + \Gamma) = R (\phi_i - \phi_1) = R \xi_i.
\end{align}
Thus, we have $\xi'_i = R \xi_i$.

\textit{Step 2:} Rotation to Align Second Amino Acid Along Positive $x$-Axis: since $\Vert \xi_2 \Vert = \Vert \xi'_2 \Vert = 1$, we have:
\begin{align}
    R_1 \xi_2 &= e_{x}, \\
    R'_1 \xi'_2 &= e_{x},
\end{align}
where $e_{x} = [1, 0, 0]^\top$. Since $\xi'_2 = R \xi_2$, we have:
\begin{equation}
    R'_1 R \xi_2 = e_{x}.
\end{equation}
Let $R'_1 = R_1 R^{-1}$, then:
\begin{equation}
    R'_1 \phi'_i = R_1 R^{-1} R \phi_i = R_1 \phi_i.
\end{equation}
\textit{Step 3:} Rotation Around $x$-Axis to Place Third Amino Acid in $xy$-Plane. Find rotation matrices $R_2$ and $R'_2$ (rotations around the $x$-axis) such that:
\begin{align}
    R_2 R_1 \phi_3 &\in \mathrm{span}\{e_x, e_y\}, \\
    R'_2 R'_1 \phi'_3 &\in \mathrm{span}\{e_x, e_y\}.
\end{align}

Since $R'_1 d'_3 = R_1 d_3$, we have:
\begin{equation}
    R'_2 R_1 \phi_3 = R_2 R_1 \phi_3.
\end{equation}
Thus, $R'_2 = R_2$. After applying the sequence of transformations, the final coordinates are:
\begin{align}
    \tilde{\phi}_i &= R_2 R_1 \phi_i, \\
    \tilde{\phi}'_i &= R'_2 R'_1 d'_i = R_2 R_1 \phi_i = \tilde{\phi}_i.
\end{align}
Thus, $ \tilde{\phi}'_i = \tilde{\phi}_i$, proving that the transformed coordinates are invariant under any initial rotation $R$ and translation $\Gamma$. This confirms that the method achieves SE(3)-invariance.

\begin{wrapfigure}{r}{0.7\linewidth}
\vspace{-2mm}
\includegraphics[width=0.99\linewidth]{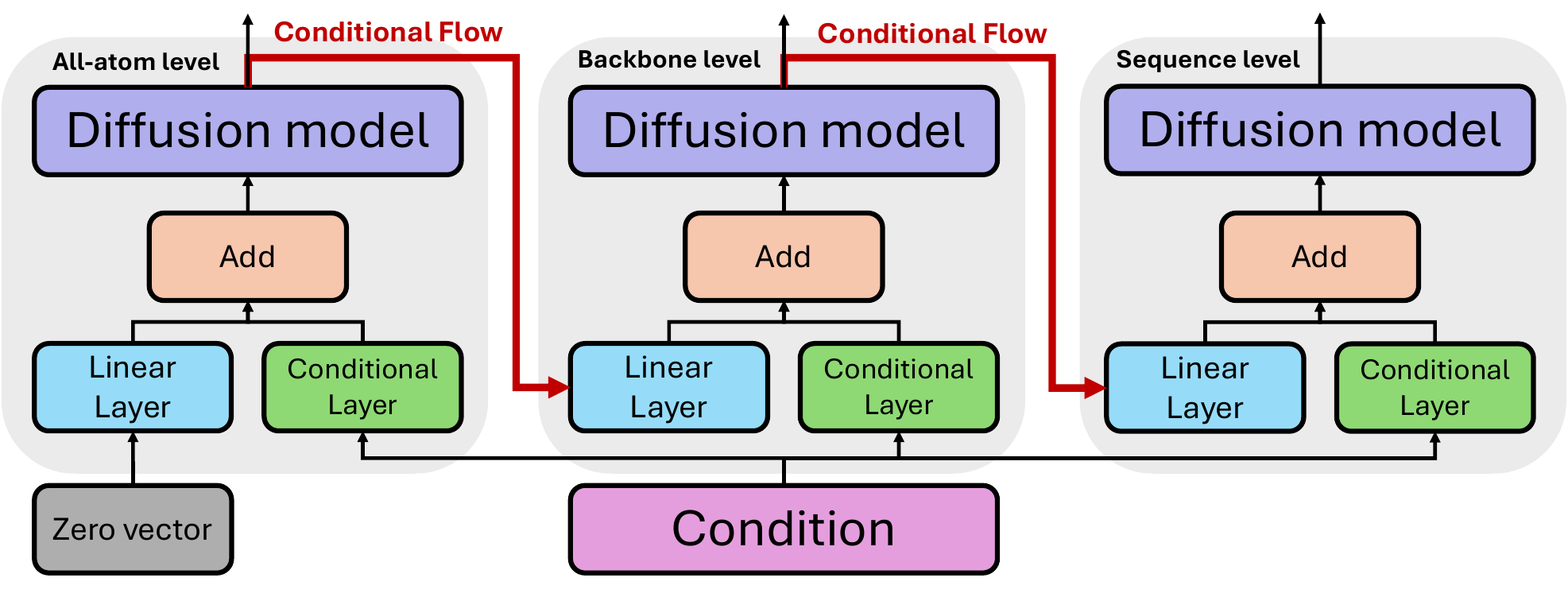} 
\caption{Consistency in the latent space.}
\vspace{-3mm}
\label{figure:condition}
\end{wrapfigure}
\vspace{1mm}
\noindent\textbf{Hierarchical Diffusion with Conditional Flow:}
To achieve control over the conditional generation of proteins at multiple levels, we employ a novel hierarchical diffusion model with a conditional flow mechanism. This design enables fine-grained manipulation of protein structure generation under specific conditions, such as targeted functional attributes. The diffusion process is split into three distinct levels: all-atom, backbone, and amino acid (sequence). Conditional information is injected from a lower level to ensure conditional consistency.


\begin{wrapfigure}{R}{0.6\textwidth}
\vspace{-6mm}
    \begin{minipage}{0.6\textwidth}
\begin{algorithm}[H]
\caption{Training Diffusion Models with Conditional Flow}
\begin{algorithmic}[1]
\While{$\text{epoch} < \text{epochs}$}
    \State Sample a random timestep $t$
    \ForAll{levels $i \in \{1, 2, 3\}$ \textbf{in parallel}}
        \If{$i = 1$}
            \State Initialize zero vector $z^{0}_{t}$
        \Else
            \State Initialize $z^{i-1}_{0}$ from ground truth data
        \EndIf
        \State Sample noise vectors
        \State Diffuse latent vectors to get $z^{i-1}_{t}$ and $z^{i}_{t-1}$
        \State Update latent vector: 
            \State \quad $z^{i}_{t} \leftarrow \epsilon^{i}(z^{i}_{t-1}; z^{i-1}_{t} W^{i}, c, \gamma_{t})$
        \State Compute loss at $i$th level
        \State Update model parameters
    \EndFor
    \State $\text{epoch} += 1$
\EndWhile
\end{algorithmic}
\label{alg: training}
\end{algorithm}
  \end{minipage}
  \end{wrapfigure}

Our conditional flow mechanism facilitates the transfer of information from lower levels (atom) to higher levels (backbone and amino acid) during the generation process. After denoising at each level, the latent representation is passed upward through a linear projection. Figure~\ref{figure:condition}
shows the conditional flow (red lines).
Specifically, for each level, the conditional flow integrates the latent vector from the lower level through a projection operation, which aligns the latent vector of the lower level to the higher level’s embedding space via a learned linear transformation. This ensures that the structural information from the previous level is preserved and effectively conditions the next level’s generation. The input at the atom level starts as a zero vector $z^{0}_{t} = 0 \in \mathbb{R}^{L \times d}$. At the higher levels, the latent vector from the previous level, after removing noise, is linearly projected and combined with the current level’s conditional embedding and time step embedding to ensure that the generative process is guided by both the condition and the structural information from the lower levels. The update at level $i$ is given by:
\begin{equation}
    z^{i}_{t} = \epsilon^{i}(z^{i}_{t-1}; z^{i-1}_{t} W^{i}, c, \gamma_{t}),
\end{equation}
where $z^{i}_{t} \in \mathbb{R}^{L \times d}$ is the latent vector at level $i$ and time step $t$, $z^{i-1}_{t}$ is the latent vector from the previous level, $W^{i} \in \mathbb{R}^{d \times d}$ is a learned linear projection matrix, $c$ represents the conditional embedding (e.g., the protein's functional target), and $\gamma_{t}$ is the time step embedding. Denote $\epsilon^{i}$ as the diffusion model at level $i$, which predicts the noise added during the forward process.

\vspace{-1mm}
\noindent\textbf{Training with Teacher Forcing:} To enable efficient parallel training, we use the teacher forcing method during training. In this setup, for each level, the input $z_{0}^{i-1}$ to the conditional flow is the ground truth data from the previous level, rather than the model's own generated output. This allows us to decouple the training of the three levels, enabling them to be trained independently and in parallel. The training process for the diffusion model at each level follows the typical DDPM framework but with the conditional flow incorporated to introduce additional control over the generative process. The training procedure is outlined in Algorithm~\ref{alg: training}.

\vspace{-3mm}
\subsection{Evaluation of Conditional Consistency}
Evaluating the quality and consistency of protein generation models requires a well-defined framework, particularly in the context of conditional generation. In this section, we define the theoretical basis for assessing conditional consistency in multi-class generation tasks and propose a novel framework to assess the suitability of different evaluations.



Denote $\{{C}^1, {C}^2, \dots, {C}^K\}$ as a set of target classes, where each class ${C}^k$ corresponds to an independent and mutually exclusive category (e.g., different protein functions or classes).
Let $x$ represent a data sample, and $d(x, {C}_k)$ be a conditional consistency metric that measures the consistency between a sample $x$ and the target class ${C}_k$. Given a model exhibiting strong conditional consistency, it should generate samples such that as we progress through a sequence of generated samples $\{x^i | i=0,1,2,\cdots, \infty\}$ ordered by increasing quality (i.e., this sequence is assumed to exist with each sample $x^i$ becoming more consistent with the target class ${C}_k$ as $i$ increases), the consistency distance between each sample and samples in $C_k$ should decrease. Mathematically, a good evaluation metric $d$ satisfies:
\begin{equation}
\lim_{i \to \infty} d(x^i, {C}^k) \rightarrow 0.
\end{equation}
It implies that as the sample quality improves, the consistency to the correct target class decreases asymptotically towards zero. We can further derive the following theorem.

\noindent{\textbf{Theorem: }}$\exists N \in \mathbb{N}^{+}, \forall i > N, \, d(x^i, {C}^k) < \min\limits_{j\neq k} d(x^i, {C}^j)$
where $C^j$ is any other class.

\noindent{\textit{Proof}}: see Appendix~\ref{sec:mmd}.


\begin{wrapfigure}{r}{0.4\linewidth}
\vspace{-2mm}
\includegraphics[width=0.85\linewidth]{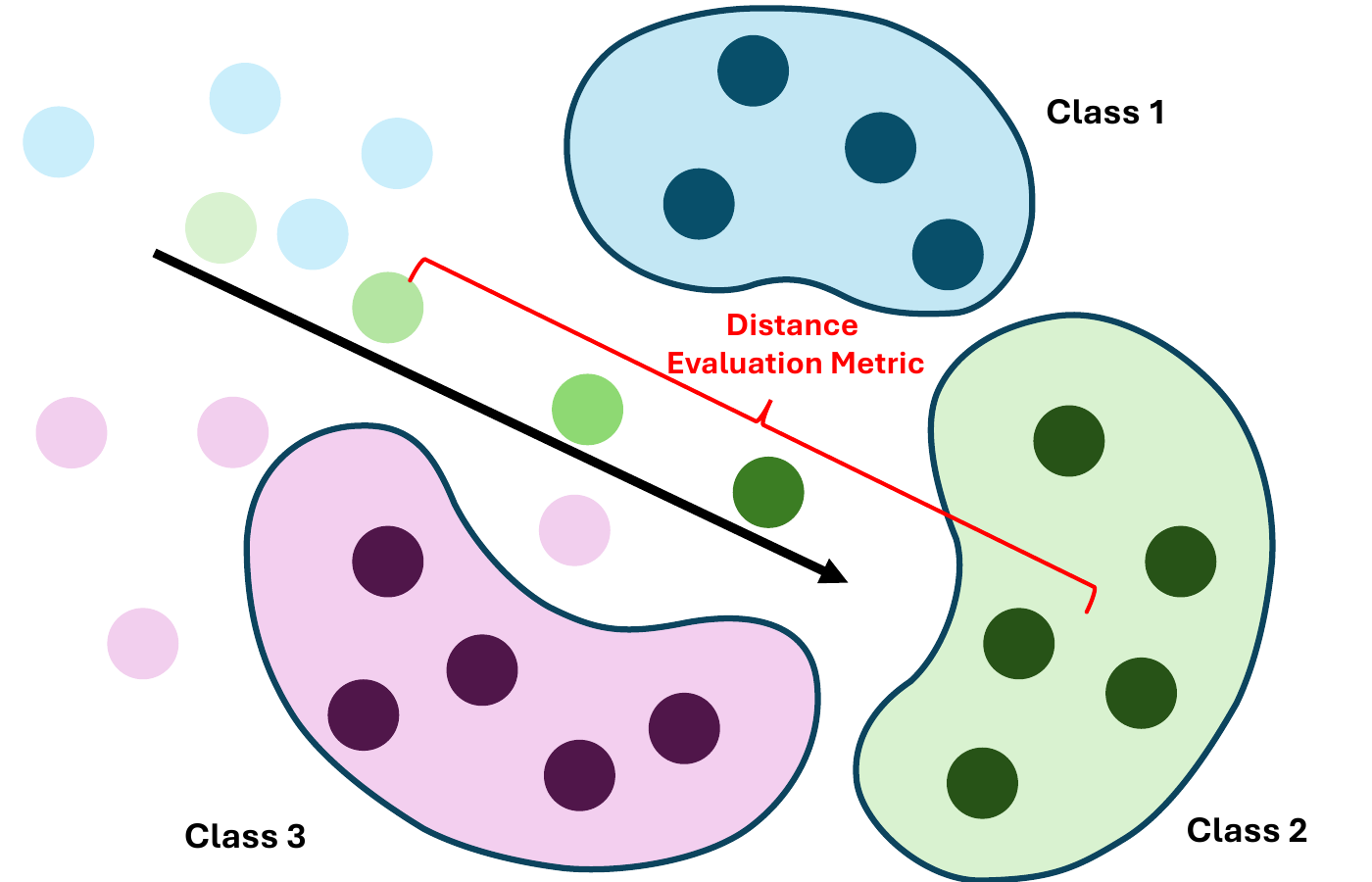}
\vspace{-2mm}
\caption{Consistency in the latent space}
\vspace{-6mm}
\label{figure:distance}
\end{wrapfigure}

Given that test samples exhibit strong conditional consistency, the theorem suggests that if we measure $d(\cdot)$ between test samples and all target classes, the majority will be classified into the correct target class ${C}^k$. However, relying solely on spatial distance may be too rigid for general evaluation, especially in conditional settings. In Figure~\ref{figure:distance}, the green points represent generated samples, and darker shades indicate better sample quality.  A well-defined metric should indicate that the green points are closer to their correct target class (i.e., Class 2) rather than the blue or pink classes.

Besides the accuracy (which class the generated belongs to), Mean Reciprocal Rank (MRR) and Normalized Mean Rank (NMR) are widely used to assess how well the evaluation metric ranks generated samples based on their correct target classes. Specifically:
\begin{equation}
\vspace{-1mm}
\text{MRR} = \frac{1}{|Q|} \sum_{i=1}^{|Q|} \frac{1}{\text{rank}_i}, \\
\text{NMR} = \frac{1}{|Q|} \sum_{i=1}^{|Q|} \frac{\text{rank}_i - 1}{N - 1}.
\end{equation}
where $Q$ is the set of test queries, and $\text{rank}_i$ is the rank of the correct target class for the $i$-th test query. These metrics, in combination with accuracy, provide a more comprehensive evaluation framework for assessing conditional consistency evaluation metrics in generative models. 

In this paper, we propose \textit{Protein-MMD}, a new evaluation metric that calculates the Maximum Mean Discrepancy (MMD) based on protein embeddings. Specifically, both real and generated protein sequences are encoded using the ESM2 language model \citep{lin2023evolutionary}, which provides biologically informed embeddings. ESM2 was chosen due to its ability to capture both structural and functional properties of proteins, thanks to its pretraining on a large protein corpus. This makes ESM2 particularly effective for evaluating distributional and functional similarities between real and generated proteins, aligning with the goals of \textit{de novo} protein design:
\begin{equation}
    \text{Protein-MMD}(p_r, p_g) = \left\| \frac{1}{n} \sum_{i=1}^{n} \varphi(x_i) - \frac{1}{m} \sum_{j=1}^{m} \varphi(y_j) \right\|_{\mathcal{H}}^2,
\end{equation}
where $\varphi(\cdot)$ denotes the embeddings extracted from the language model. These embeddings represent both sequence and functional information, making them particularly well-suited for comparing real and generated protein distributions.

\begin{wraptable}{r}{0.5\textwidth}
\vspace{-4mm}
\centering
\caption{Evaluation on the EC dataset.}
\begin{tabular}{@{}lccc@{}}
\toprule
\textbf{Metric}        & \textbf{Accuracy} $\uparrow$ & \textbf{MRR} $\uparrow$ & \textbf{NMR} $\downarrow$ \\ \midrule
MMD                   & 0.0687         & 0.3101        & 0.5506         \\
Protein-FID           & 0.2988         & 0.4825        & 0.3920         \\
\textbf{Protein-MMD}  & \textbf{0.5487} & \textbf{0.6629} & \textbf{0.2524} \\ \bottomrule
\end{tabular}
\vspace{-3mm}
\label{table:metrics}
\end{wraptable}
To validate the effectiveness of Protein-MMD and other metrics, we apply the evaluation metrics on the Enzyme Commission (EC) dataset, which categorizes proteins based on the reactions they catalyze using EC numbers. We focus on seven classes from the first EC number category for our conditional generation task. In Table~\ref{table:metrics}, we compare three evaluation metrics: MMD (considering only sequence statistics as presented in \citep{kucera2022conditional}), Protein-FID (using ESM2 in place of Inception for protein generation), and Protein-MMD. All metrics are used to compute the Accuracy, Mean Reciprocal Rank (MRR), and Normalized Mean Rank (NMR) scores to evaluate how it performs in evaluating the conditional consistency. As observed, \textit{Protein-MMD} outperforms both MMD and Protein-FID across all evaluation metrics. The higher accuracy and MRR scores indicate that \textit{Protein-MMD} better captures the conditional consistency of the proteins in the test set. The lower NMR score further demonstrates that \textit{Protein-MMD} ranks the correct target class higher in comparison to other metrics, validating its effectiveness in conditional protein generation tasks. While \textit{Protein-MMD} proves to be the most effective metric according to our framework, we acknowledge the widespread use of FID in generative modeling tasks. Therefore, we will continue to report Protein-FID results alongside Protein-MMD in subsequent experiments.

\vspace{-3mm}
\section{Experiments}
\vspace{-1mm}
\subsection{Experimental Setup}
\vspace{-1mm}
We compared our model against several baselines, each representing distinct approaches to protein generation. ProteoGAN \citep{kucera2022conditional} is a GAN-based method, while ESM2 \citep{lin2023evolutionary} and ProstT5 \citep{heinzinger2023prostt5} are Transformer-based language models specifically designed for protein sequence modeling. ProteinMPNN \citep{dauparas2022robust} and LatentDiff \citep{fu2024latent}, on the other hand, are graph-based models, with LatentDiff also incorporating a diffusion-based framework, specifically using a latent diffusion approach. For each model, we evaluate the performance using both diversity metrics (TM-score, RMSD, and Seq.ID) and conditional consistency metrics (Protein-MMD and Protein-FID). Higher RMSD, lower TM-score, and lower Seq.ID indicate higher diversity, while lower Protein-MMD and Protein-FID values signify higher conditional consistency between the generated and real protein distributions. More detailed settings can be found in Appendix~\ref{exp:setting}.

\begin{table*}[h]
\centering
\vspace{-2mm}
\caption{Results on EC and GO datasets.}
\begin{adjustbox}{width=\linewidth}
\begin{tabular}{lccccc|ccccccc}
\toprule
\multirow{2}{*}{} & \multicolumn{5}{c}{EC Dataset} & \multicolumn{5}{c}{GO Dataset} \\  
\cmidrule(lr){2-6} \cmidrule(lr){7-11}
 & \multicolumn{3}{c}{Diversity} & \multicolumn{2}{c}{Conditional Consistency} & \multicolumn{3}{c}{Diversity} & \multicolumn{2}{c}{Conditional Consistency} \\
 & TM-score$\downarrow$ & RMSD$\uparrow$ & Seq.ID$\downarrow$ & Protein-MMD$\downarrow$ & Protein-FID$\downarrow$ & TM-score$\downarrow$ & RMSD$\uparrow$ & Seq.ID$\downarrow$ & Protein-MMD$\downarrow$ & Protein-FID$\downarrow$ \\
\midrule
ProteGAN & 0.26 & \underline{5.35} & 6.71 & 13.99 & 260.31 & 0.23 & 5.96 & \textbf{6.33} & \textbf{10.89} & \textbf{256.31} \\
ESM2 & 0.29 & 4.25 & \textbf{6.57} & \underline{13.35} & \underline{238.46} & \underline{0.22} & \textbf{7.33} & \underline{6.39} & 11.86 & 290.31 \\
ProstT5 & 0.28 & 4.25 & \underline{6.61} & 13.76 & 248.32 & 0.26 & 6.81 & 6.73 & 11.93 & 292.58 \\
ProteinMPNN & \textbf{0.24} & 4.24 & 67.43 & 22.31 & 587.72 & \textbf{0.14} & \underline{7.10} & 77.96 & 15.94 & 410.43 & \\
LatentDiff & 0.37 & 2.73 & 7.67 & 13.43 & 256.75 & 0.31 & 4.26 & 7.37 & 12.66 & 346.40 \\
\hline
Ours(128) & \textbf{0.24} & 4.7 & 7.56 & 13.74 & 250.2 & \multicolumn{5}{c}{-----} \\
Ours(256) & 0.27 & 4.40 & 6.88 & 13.67 & 248.31 & \multicolumn{5}{c}{-----} \\
Ours(512) & \underline{0.25} & \textbf{5.39} & 6.79 & \textbf{13.28} & \textbf{237.46} & 0.26 & 6.09 & 7.13 & \underline{11.67} & \underline{284.65} \\
\bottomrule
\multicolumn{11}{r}{The best performance for each metric is indicated in \textbf{bold}, while the second-best performance is \underline{underlined}.}
\end{tabular}
\end{adjustbox}
\vspace{-6mm}
\label{table: results}
\end{table*}

\vspace{-2mm}
\subsection{Results and Analysis}
Table~\ref{table: results} presents the results of our model and the baselines on two datasets. Our model achieves the best performance in terms of most metrics on the EC dataset, indicating superior conditional consistency and diversity in generating proteins that adhere closely to the specified enzyme classes. On the EC dataset, our model (with sequence length 512) achieves the lowest Protein-MMD and Protein-FID scores, demonstrating effective modeling of the distributional and functional similarities between generated and real proteins. The RMSD and TM-score metrics indicate that our model generates structurally diverse proteins, with the highest RMSD and among the 2nd-lowest TM-scores, suggesting less topological similarity to templates. The sequence identity (Seq.ID) is also low, indicating higher sequence diversity. 
For the GO dataset, our model also performs competitively. However, in terms of conditional consistency metrics (Protein-MMD and Protein-FID), our model ranks second, with ESM2 achieving the best Protein-MMD score and ProteoGAN achieving the best Protein-FID score. This suggests that our model generates diverse protein structures.

\begin{wraptable}{r}{0.55\textwidth}
\centering
\vspace{-4mm}
\caption{Ablation study.}
\begin{tabular}{@{}lccc@{}}
\toprule
Method & Protein-MMD$\downarrow$ & Protein-FID$\downarrow$ \\
\midrule
All & \textbf{13.50} & \textbf{241.82} \\
Removed backbone level & 13.73 & 249.14 \\
Removed all-atom level & 13.94 & 251.83 \\
Removed both & 14.06 & 255.15 \\
\bottomrule
\end{tabular}
\vspace{-3mm}
\label{table: ablation}
\end{wraptable}
\noindent\textbf{Ablation Study.} To investigate the effectiveness of each of the three levels (amino acid, backbone, all-atom), we conducted an ablation study in the experiment. Specifically, the variant of our model removes either a specific level (the backbone or all-atom) or both two levels. Then we examine the performance of the conditional consistency metrics. Note that we can not remove the amino acid level because the amino acid is required for evaluation.  The ablation study is conducted on the EC dataset. As shown in Table~\ref{table: ablation},  if we remove any level (i.e., backbone and all-atom level) or both two levels, the performance will drop. It verifies the necessity of our multi-level conditional diffusion.  

\begin{wrapfigure}{r}{0.35\linewidth}
\vspace{-5mm}
\includegraphics[width=\linewidth]{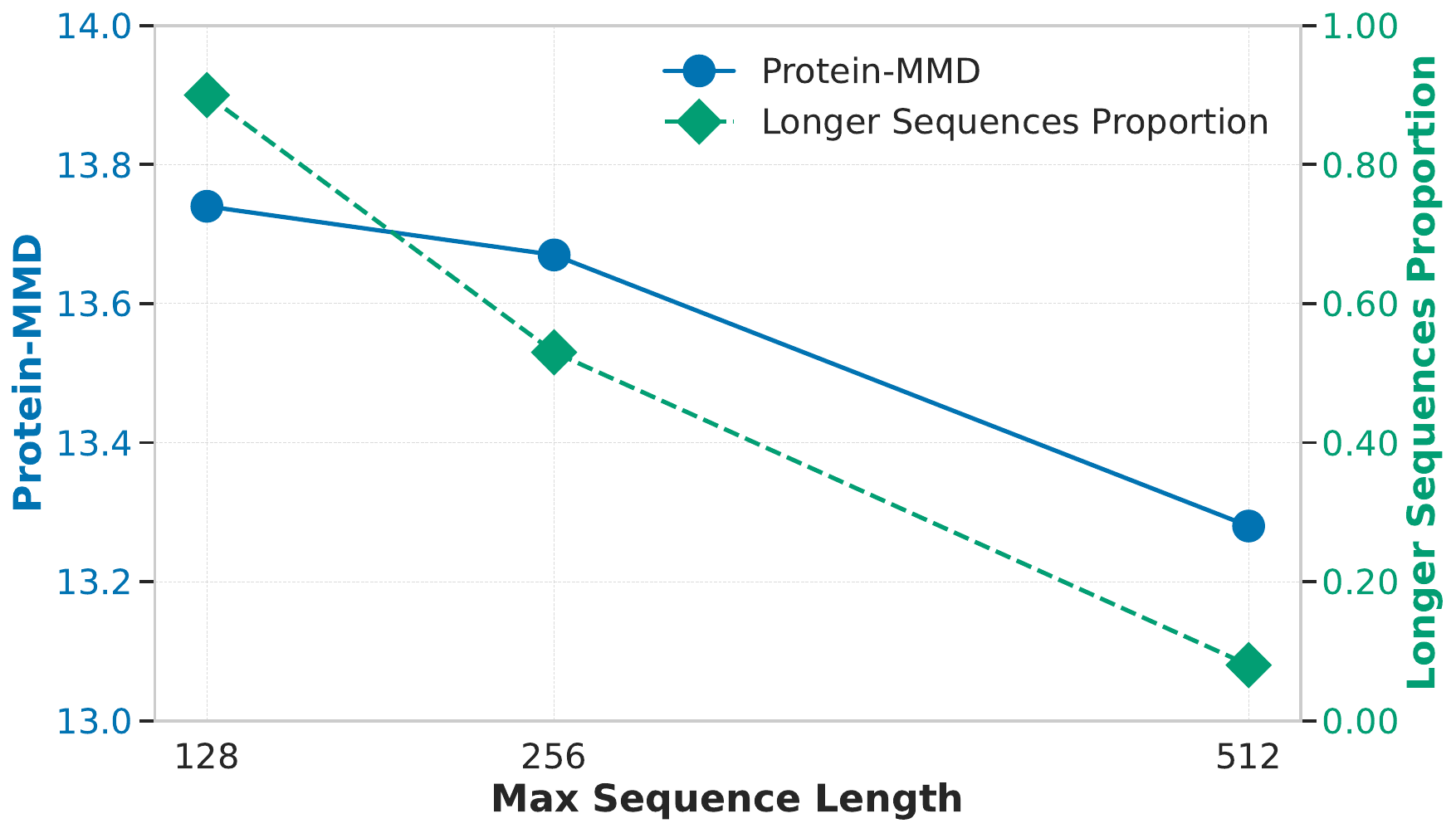}
\vspace{-3mm}
\caption{Consistency.}
\vspace{-4mm}
\label{figure: len}
\end{wrapfigure}
\noindent\textbf{Impact of Maximum Sequence Length.} In previous studies on protein \textit{de novo} design, existing works usually employ a maximum sequence length of 128~\citep{fu2024latent}. However, through our experiments, we observed that for conditional generation tasks, shorter sequence lengths fail to fully leverage the conditional information, which in turn results in lower conditional consistency metrics. To address this, we constructed models with three different maximum sequence lengths: 128, 256, and 512, and investigated the impact of maximum length on the model’s ability to maintain conditional consistency.

\begin{wraptable}{r}{0.4\textwidth}
\centering
\vspace{-4mm}
\caption{Comparison with ProteoGAN.}
\vspace{-1mm}
\begin{tabular}{@{}lccc@{}}
\toprule
Method & IoU\textsubscript{mean}$\uparrow$ & IoU\textsubscript{max}$\uparrow$ \\
\midrule
ProteoGAN & \textbf{0.2181} & 0.4706 \\
Ours (512) & 0.2088 & \textbf{0.5833} \\
\bottomrule
\end{tabular}
\vspace{-3mm}
\label{table: iou_results}
\end{wraptable}
As shown in Figure~\ref{figure: len}, we observe a positive correlation between the Protein-MMD metric, which reflects conditional consistency, and the proportion of training data samples exceeding the current maximum sequence length. This indicates that longer sequences help the model better incorporate condition information during generation. Moreover, the results in Table~\ref{table: results} for our method with different lengths reveal that the maximum sequence length does not influence the model's performance on diversity metrics, which are independent of the quality of condition-guided generation.  These findings underscore the importance of maximum sequence length in enhancing conditional consistency, offering valuable insights for the design of future protein conditional generation models.

\begin{wrapfigure}{R}{0.5\linewidth}
\vspace{-2mm}
\includegraphics[width=0.99\linewidth]{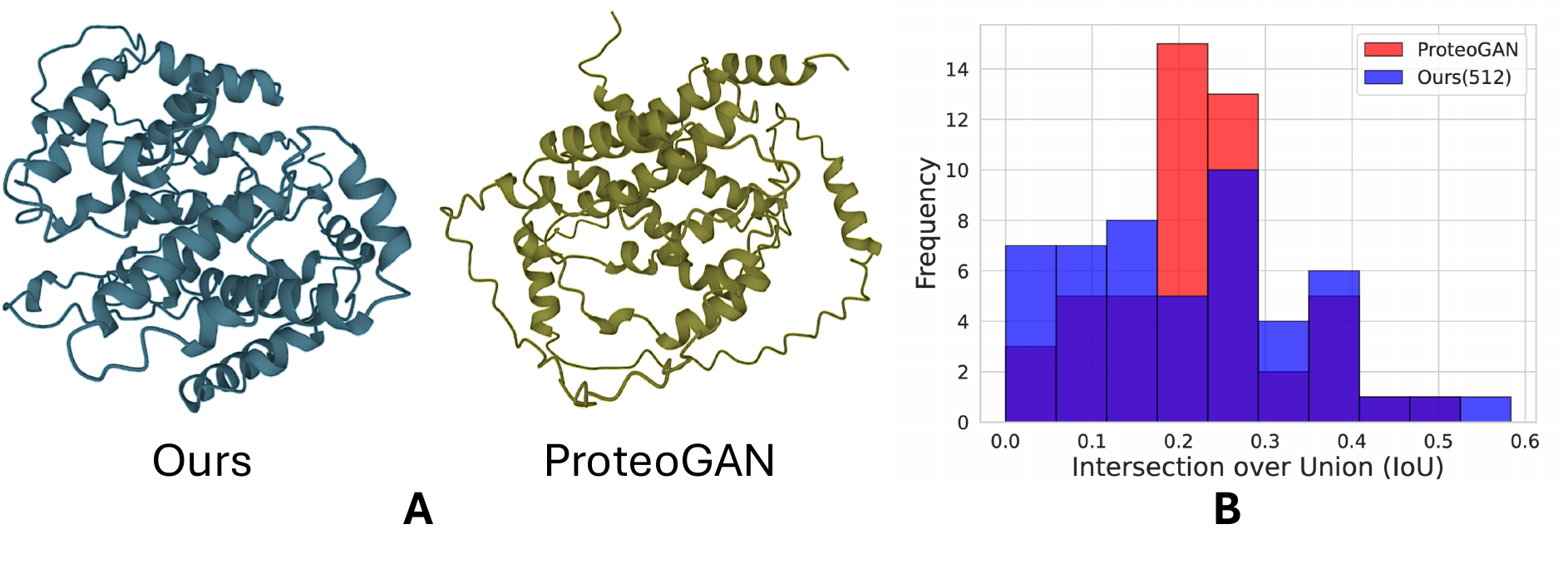}
\vspace{-3mm}
\caption{\textbf{A} shows the two highest generated protein results of Ours and ProteoGAN in terms of the IoU indicator. \textbf{B} shows the statistical frequency histogram.}
\vspace{-1mm}
\label{figure:case_study}
\end{wrapfigure}
\noindent\textbf{Case Study.} To further demonstrate the superiority of our model on the GO dataset, particularly regarding conditional consistency, we conducted a fine-grained case study comparing our method with the best baseline ProteoGAN. We utilized an \textit{in-silico} evaluation to perform a fine-grained analysis of the generated protein sequences. By employing a trained ESM-MLP classifier on the GO dataset, we assessed each generated protein's adherence to the specified GO terms using the Intersection over Union (IoU)~\citep{DBLP:conf/cvpr/RezatofighiTGS019}.
As shown in Table~\ref{table: iou_results}, our method exhibits a lower average IoU\textsubscript{mean} compared to ProteoGAN, aligning with earlier results in Table~\ref{table: results}. However, it achieves a higher IoU\textsubscript{max}, indicating a greater potential for generating high-quality samples that closely match the desired GO annotations. Figure~\ref{figure:case_study} illustrates the distribution of IoU scores. While ProteoGAN's samples are concentrated around medium quality, our method generates a broader range of samples, including those with higher IoU scores.  This suggests that our model, despite a lower average performance, is more capable of producing proteins with superior conditional consistency.

\section{Related Works}
\textit{De novo} protein design methods are dedicated to identify novel proteins with the desired structure and function properties~\citep{watson2023novo, huang2016coming, DBLP:conf/iclr/FreyBZKLHWRBCLG24,DBLP:conf/iclr/MaoZSSW0S24,DBLP:conf/iclr/KomorowskaMDVLJ24,DBLP:conf/iclr/GaoSLL0024}. Recent advancements in machine learning have enabled a generative model to accelerate
key steps in the discovery of novel molecular structures and drug design~\citep{DBLP:conf/nips/GaoQTJRLLML23,DBLP:conf/nips/LuWZRLZ22,DBLP:conf/nips/PeiGWZX000L023,DBLP:conf/icml/LiuLUMJ22}. A prior step of generate models to the representation of proteins~\citep{DBLP:conf/iclr/ZhangXJCLD023,DBLP:conf/iclr/ZhouFZBY23,DBLP:conf/icml/GongKLC0D24}. The majority of representation learning for protein is to represent a protein as a sequence of amino acids~\citep{DBLP:conf/iclr/ChenGDCW0WEDK23,DBLP:conf/iclr/MoretaRAHTH22,DBLP:conf/iclr/LeeY0K24,DBLP:conf/iclr/FanW0K23}. Considering the spatial information is important to the property of a protein, many works resort to a graph model for a comprehensive presentation with the structure information~\citep{DBLP:conf/iclr/IngrahamGBJ19,DBLP:conf/aaai/HuangLWSLZLGZL24}. In general, each node on the graph is an amino acid and the edge is decided by the distance between two nodes~\cite{DBLP:conf/kdd/AykentX22,DBLP:conf/kdd/XiaK21,DBLP:conf/iclr/HladisLFT23}. Despite the power of graph models, the relation information in a 3-dimensional space captures the multi-level structure such as the angle between two edges. A line of research works explore the protein structure in 3D space~\cite{DBLP:conf/iclr/HermosillaSLFVK21,DBLP:conf/iclr/Huang0ZZZZCW0Y24}. Recently, large language models (LLMs) have also been introduced to model the sequence~\cite{DBLP:conf/nips/MeierRVLSR21,lin2023evolutionary} inspired by the success of natural language processing.

Capitalizing on the power of generative
models such as Generative Adversarial Networks (GANs) and diffusion models, deep generative modeling has shown its potential for fast generation of new and viable protein structures. \cite{DBLP:conf/nips/AnandH18} has applied GANs to the task of generating protein structures by encoding protein structures in terms of pairwise distances on the protein backbone. Diffusion models~\cite{song2021denoising,jiang2023successfully,gao2024image,xiao2025lightcache} have emerged as a powerful tool for graph-structured diffusion processes~\cite{DBLP:conf/icml/KlarnerRMDT24}. FrameDiff has been proposed for monomer backbone generation and it can generate designable monomers up to 500 amino acids~\cite{DBLP:conf/icml/YimTBMDBJ23}. NOS is another diffusion model that generates protein sequences with high likelihood by taking many alternating steps in the continuous
latent space of the model~\cite{DBLP:conf/nips/GruverSFRHLRCW23}.

\section{Conclusions}
In this paper, we introduce a novel multi-level conditional generative diffusion model that integrates sequence-based and structure-based information for efficient end-to-end protein design. Our model incorporates a 3D rotation-invariant preprocessing step to maintain SE(3)-invariance. To address the limitation of the existing evaluations, we propose a novel metric to evaluate conditional consistency.

 \clearpage
\bibliography{example_paper}
\bibliographystyle{IEEEtran}

\appendix
\section{Appendices}
\vspace{-2mm}

\begin{figure*}[t]
\centering
\includegraphics[width=\linewidth]{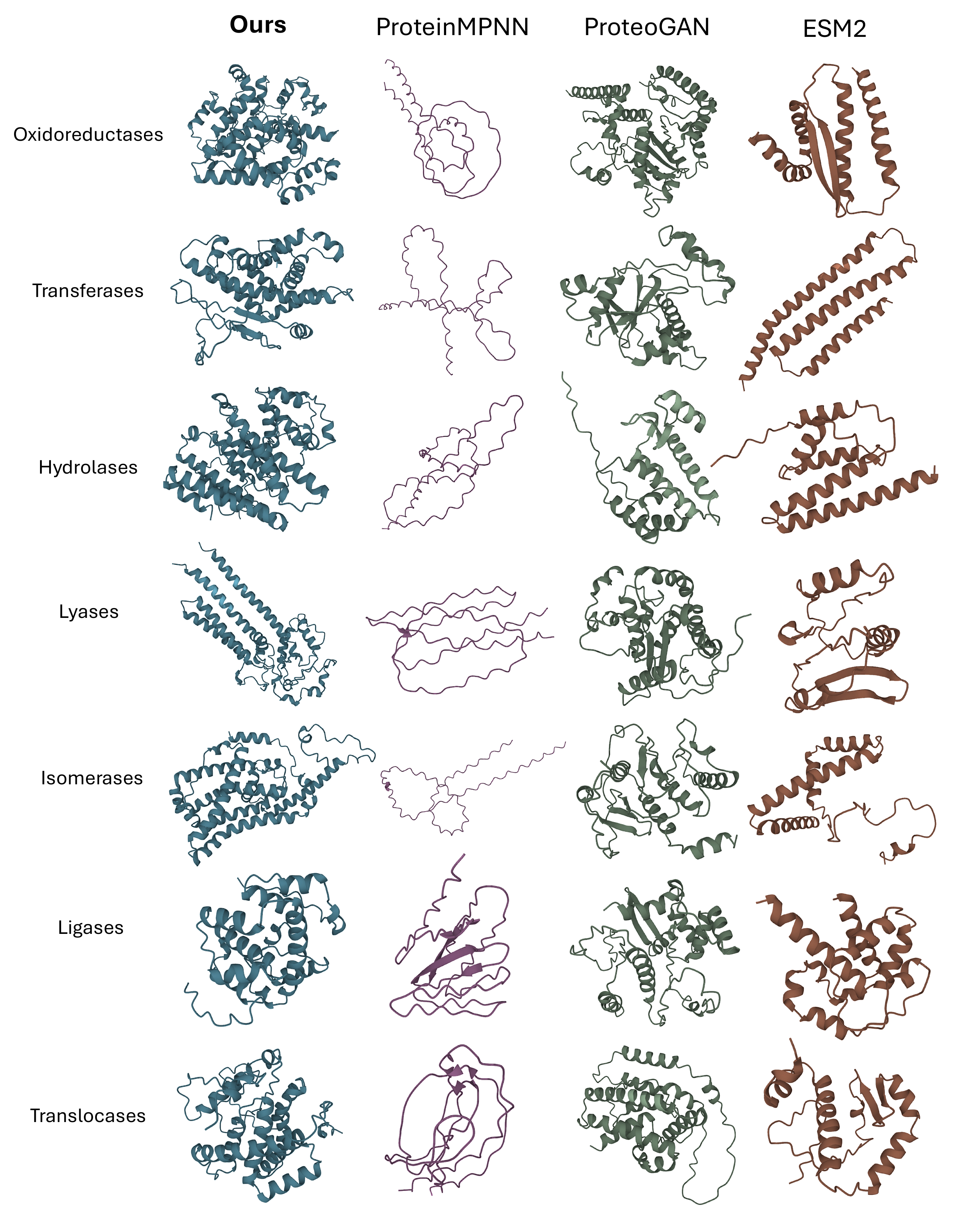}
\vspace{-2mm}
\caption{Protein Visualization Comparison on EC Dataset (Ours vs. ProteinMPNN, ProteoGAN, and ESM2).}
\label{figure: vis-all}
\vspace{-2mm}
\end{figure*}
\vspace{1mm}

\subsection{Experiment settings}\label{exp:setting}
To verify the effectiveness of our proposed multi-level conditional diffusion model, we conducted comprehensive experiments on two standard datasets: the Enzyme Commission (EC) dataset and the Gene Ontology (GO) dataset. The EC dataset categorizes proteins based on the biochemical reactions they catalyze, while the GO dataset classifies proteins according to their associated biological processes, cellular components, and molecular functions. These datasets provide a robust benchmark for assessing both the diversity and conditional consistency of generated protein sequences.

Our model leverages the \texttt{esm2\_t33\_650M\_UR50D} model from ESM2 \citep{lin2023evolutionary} as the amino acid-level encoder. To construct the Protein Variational Auto-encoder model, we set the latent dimension to 384, and the decoder is composed of 8 Transformer \citep{vaswani2017attention} decoder blocks, each equipped with an 8-head self-attention mechanism. The Protein Variational Auto-encoder model is trained with a learning rate of $10 ^ {-4}$, using a combination of mean squared error (MSE) and cross-entropy as the loss functions. To regulate the latent vector distribution, we apply a KL divergence loss with a weight of $10^{-5}$. We experimented with 128, 256, and 512 as the maximum sequence lengths. For the diffusion model, we modify the DiT-B architecture from DiT \citep{peebles2023scalable}, which consists of 12 DiT blocks and uses a hidden size of 768. The DiT model is trained from scratch with a learning rate of $10 ^ {-4}$ and includes a weight decay of $10 ^ {-5}$.
\subsection{Protein-MMD}\label{sec:mmd}
\noindent{\textbf{Theorem: }}$\exists N \in \mathbb{N}^{+}, \forall i > N, \, d(x^i, {C}^k) < \min\limits_{j\neq k} d(x^i, {C}^j)$
where $C^j$ is any other class.

\noindent{\textit{Proof}}: Assume that there exists a class ${C}^j (j\neq k)$ such that $d(x^i, {C}^j) \leq d(x^i, {C}^i)$ for $i >N$.
Since ${C}^k$ is defined as the correct target class and the quality of the generated sample improves with $i \to \infty$, the consistency $d(x_n^i, {C}^k)$ should approach zero. If $d(x^i, {C}^j) \leq d(x^i, {C}^i)$, we have $\lim\limits_{n \to \infty} d(x^i, {C}^j) = 0$. It contradicts the assumption that $C^k$ is the correct class for the generated data. Therefore, the assumption is false.



\noindent\textbf{Visualization.}
In Figure~\ref{figure: vis-all} (Appendix), we present visualizations of proteins conditionally generated by our method and other baselines on the EC dataset. Specifically, we generate proteins with 7 different functions (e.g., Oxidoreductases). Compared with the baselines, our method can generate discriminative proteins given the same input. By modeling the hierarchical relation at different levels, our method can generate foldable and functional sequences in 3D space.
\vspace{-2mm}

\end{document}